%% file: event_gan_sim.tex
\documentclass[10pt,twocolumn,letterpaper]{article}

\usepackage{cvpr}
\usepackage{times}
\usepackage{epsfig}
\usepackage{graphicx}
\usepackage{amsmath}
\usepackage{amssymb}
\usepackage{caption}
\usepackage{subcaption}
\usepackage{multirow}
\usepackage{floatrow}
\newfloatcommand{capbtabbox}{table}[][\FBwidth]

\newcommand\blfootnote[1]{%
  \begingroup
  \renewcommand\thefootnote{}\footnote{#1}%
  \addtocounter{footnote}{-1}%
  \endgroup
}

\newcommand{\loss}{\mathcal{L}}
\newcommand{\x}{\mathbf{x}}

\usepackage[pagebackref=true,breaklinks=true,letterpaper=true,colorlinks,bookmarks=false]{hyperref}

\cvprfinalcopy 

\def\cvprPaperID{****} 

\ifcvprfinal\fi
\begin{document}

\title{EventGAN: Leveraging Large Scale Image Datasets for Event Cameras}

\author{Alex Zihao Zhu \and Ziyun Wang \and Kaung Khant \and Kostas Daniilidis\\
University of Pennsylvania\\
{\tt\small \{alexzhu, ziyunw, khantk, kostas\}@seas.upenn.edu}
}

\maketitle

\begin{abstract}
Event cameras provide a number of benefits over traditional cameras, such as the ability to track incredibly fast motions, high dynamic range, and low power consumption. However, their application into computer vision problems, many of which are primarily dominated by deep learning solutions, has been limited by the lack of labeled training data for events. In this work, we propose a method which leverages the existing labeled data for images by simulating events from a pair of temporal image frames, using a convolutional neural network. We train this network on pairs of images and events, using an adversarial discriminator loss and a pair of cycle consistency losses. The cycle consistency losses utilize a pair of pre-trained self-supervised networks which perform optical flow estimation and image reconstruction from events, and constrain our network to generate events which result in accurate outputs from both of these networks. Trained fully end to end, our network learns a generative model for events from images without the need for accurate modeling of the motion in the scene, exhibited by modeling based methods, while also implicitly modeling event noise. Using this simulator, we train a pair of downstream networks on object detection and 2D human pose estimation from events, using simulated data from large scale image datasets, and demonstrate the networks' abilities to generalize to datasets with real events.
\end{abstract}

\blfootnote{The code for this project can be found at: \url{https://github.com/alexzzhu/EventGAN}.}
\input{tex/intro.tex}
\input{tex/pipeline_fig.tex}
\input{tex/related.tex}
\input{tex/method.tex}
\input{tex/gan_outputs.tex}
\input{tex/detection_qualitative.tex}
\input{tex/experiments.tex}

\input{tex/results.tex}
\input{tex/conclusions.tex}

\section{Acknowledgements}
Thanks to Tobi Delbruck and the team at iniLabs and iniVation for providing and supporting the DAVIS-346b cameras. This work was supported in part by the Semiconductor Research Corporation (SRC) and DARPA. We also gratefully appreciate support through the following grants: NSF-IIP-1439681 (I/UCRC), NSF-IIS-1703319, NSF MRI 1626008, ARL RCTA W911NF-10-2-0016, ONR N00014-17-1-2093, ARL DCIST CRA W911NF-17-2-0181, the Amazon Research Award, the AWS Cloud Credits for Research Program and Honda Research Institute.

{\small
\bibliographystyle{ieee_fullname}
\bibliography{bib}
}

\end{document}

%% file: tex/intro.tex
\section{Introduction}
Deep learning has led a revolution for many computer vision tasks which had been considered incredibly challenging. The ability to leverage immense amounts of data to train neural networks has resulted in significant improvements in performance for many tasks. As a vision modality, event cameras have a lot to gain from deep learning. By combining the neural networks with the advantages of event cameras, we stand to be able to extend the operating volume of speeds and lighting conditions significantly beyond that which is achievable by traditional cameras. 

However, these networks for events are limited by the amount of labeled training data available, due to the camera's relative infancy and the cost of acquiring accurate ground truth labels. While some works have been able to bypass this issue with self-supervised approaches~\cite{zhu2018ev, rebecq2019events, ye2019unsupervised}, some problems, such as detection and classification, cannot currently be solved without a large corpus of labeled training data. In this work, we focus on an alternative to costly data labelling, by leveraging the large set of labeled image datasets via image to event simulation. 

The highest fidelity event camera simulators today \cite{rebecq2018esim, mueggler2017event, li2018interiornet} all operate with a similar framework, by simulating optical flow in the image either through 3D camera motion, or a parametrized warping (e.g. affine) of the image, in order to precisely track the generation of events as each point in the image moves to a new pixel. However, these scenarios either require simulation of the full 3D scene, or severely constrain the motion in the image. In addition, modeling event noise, both in terms of erroneous events and noise in the event measurements, is a challenging open problem.

In this work, we present EventGAN, a novel method for image to event simulation, where we apply a convolutional neural network as the function between images and events. By learning this function with data, our method does not require any explicit knowledge of the scene or the relationship between images and events, but is instead able to regress a realistic set of events given only images as input. In addition, our network is able to learn the noise distribution over the events, which are currently not modeled by the competing methods. Finally, our proposed method has a fast, constant time simulation which is easily parallelizable on GPUs and integrable into any modern neural network architecture, as opposed to the prior work which requires 3D simulations of the scene.

Our network is trained on a set of image and event pairs, which are directly output by event cameras such as the DAVIS~\cite{brandli2014240}. At training time, we apply an adversarial loss to align the generated events with the real events. In addition, we pre-train a pair of CNNs to perform optical flow estimation and image reconstruction from real events, and constrain our generator to produce events which allow these pre-trained networks to generate accurate outputs. In other words, we constrain the generated events to retain the motion and appearance information present in the real data.

Using this event simulation network, we train a set of downstream networks to perform object detection on cars and 2D human pose estimation, given images and labels from large scale image datasets such as KITTI~\cite{Geiger2012CVPR}, MPII~\cite{andriluka14cvpr} and Human3.6M~\cite{h36m_pami}. We then evaluate performance on these downstream tasks on real event datasets, MVSEC~\cite{zhu2018multivehicle} for car detection, and DHP19~\cite{calabrese2019dhp19} for human pose, demonstrating the generalization ability of these networks despite having mostly seen simulated data at training time. All data and code will be released at a later date.

Our main contributions can be summarized as:
\begin{itemize}
\item A novel pipeline for supervised training of deep neural networks for events, by simulating events from existing large scale image datasets and training on the simulated events and image labels.
\item A novel network, EventGAN, for event simulation from a pair of images, trained using an adversarial loss and cycle consistency losses which constrain the generator network to generate events from which pre-trained networks are able to extract accurate optical flow and image reconstructions.
\item A test dataset for car detection, with manually labeled bounding boxes for cars from the MVSEC~\cite{zhu2018multivehicle} dataset.
\item Experiments demonstrating the generalizability of the networks trained on simulated data to real event data, by training object detection and human pose networks on simulated data, and evaluating on real data.
\end{itemize}

%% file: tex/pipeline_fig.tex
\begin{figure*}[t!]
    \centering
    \includegraphics[width=0.7\linewidth]{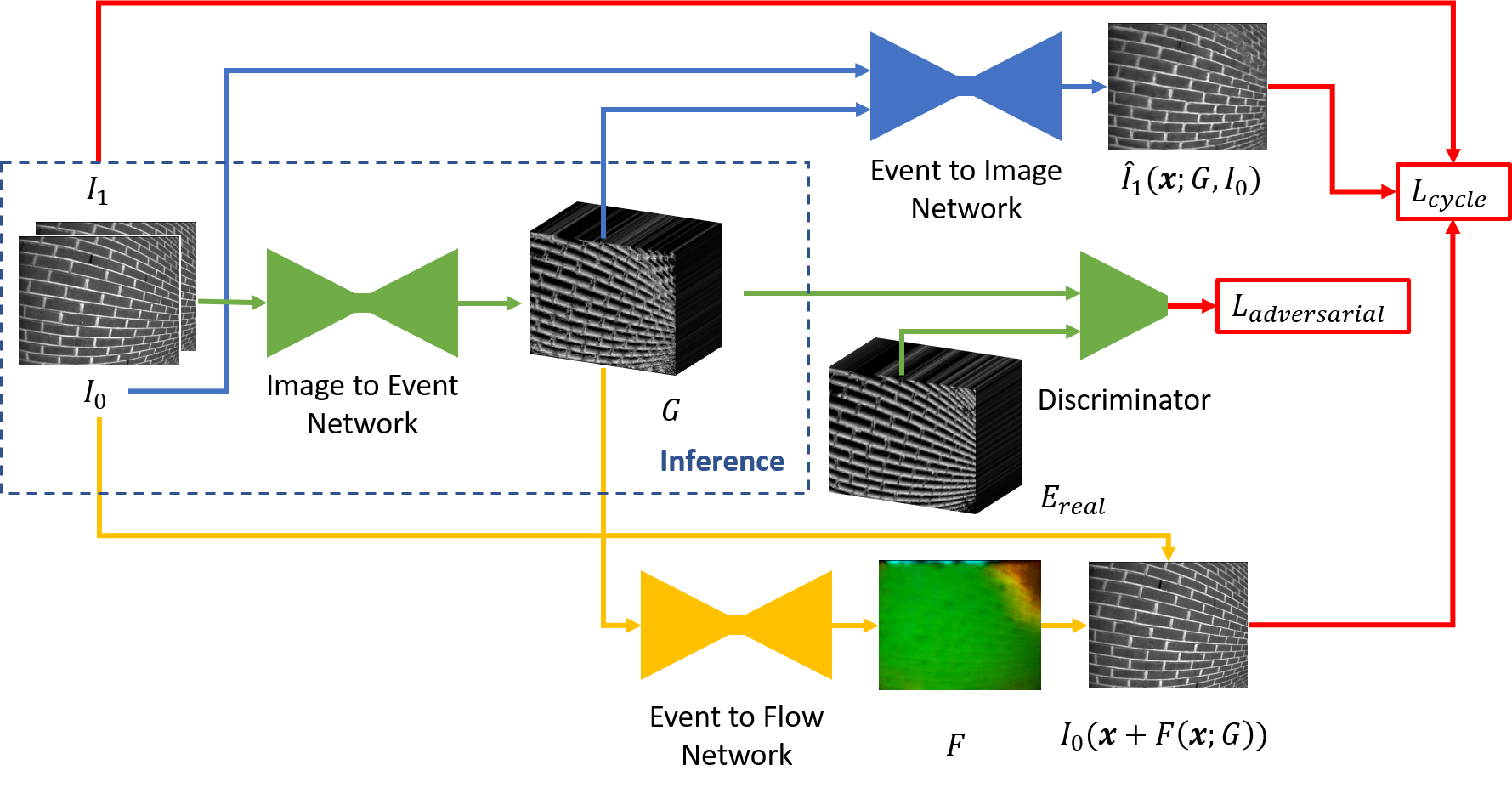}
    \caption{Overview of the EventGAN pipeline. A pair of grayscale images are passed into the generator, which predicts a corresponding event volume. This output is constrained by an adversarial loss, as well as a pair of cycle consistency losses which constrain the generated volume to encode image and flow information.}
    \label{fig:pipeline}
\end{figure*}

%% file: tex/related.tex
\section{Related Work}
\subsection{Event Simulation}
Prior works on event simulation have focused on differencing log intensity frames, in order to simulate the condition required to trigger an event:
\begin{align}
\|\log(I_{t+1}(\x))-\log(I_t(\x))\|\geq \theta
\end{align}
Earlier works by Bi et al.~\cite{bi2017pix2nvs} and Kaiser et al.~\cite{kaiser2016towards} simulating events by directly applying this equation to the log intensity difference between each pair of successive images. These methods were limited by the temporal resolution of these images, and as such could only handle relatively slow moving scenes. To improve fidelity, Rebecq et al.~\cite{rebecq2018esim}, Mueggler et al.~\cite{mueggler2017event} and Li et al.~\cite{li2018interiornet} perform full 3D simulations of a scene. This allows them to simulate images at arbitrary temporal resolution, while also having access to the optical flow within the scene, allowing for accurate event trajectories. However, these methods are limited to fully simulated scenes, or images where the motion is known (or where a simplified motion model such as an affine transform is applied). Performing 3D simulations is also a relatively expensive procedure, requiring complex rendering engines. In addition, these methods do not properly model the noise properties of the sensor. Rebecq et al.~\cite{rebecq2018esim} apply Gaussian noise to the trigger threshold, $\theta$, as an approximation, but no true model of the event noise distribution exists to our knowledge.

Our work, in contrast, runs in constant time using a CNN which is easily parallelizable and optimized for modern GPUs. The network learns both the motion information in the scene, as well as the noise distribution of the events.

\subsection{Sim2Real/Domain Adapation}
Learning from simulations and other modalities has been a rapidly growing topic, with deep learning approaches for many robotics problems in particular requiring much more training data than is practical to collect on a physical platform. However, this remains a challenging open problem, as conventional simulators often cannot perfectly model the data distribution in the real world, resulting in many methods attempting to bridge this gap~\cite{peng2018sim, james2019sim}. One popular approach to this problem in the image space is the use of Generative Adversarial Networks (GANs)~\cite{goodfellow2014generative}, which consists of a generator trained to model the data distribution of the training set, while a discriminator is trained to differentiate between outputs from the fake and real data. With particular relevance to this work, conditional GANs~\cite{mirza2014conditional, isola2017image} are able to model relationships between data distributions, while CycleGANs apply additional cycle consistency losses~\cite{zhu2017unpaired}.

A successful application of cross modality transfer is in the field of image to lidar transform. A number of recent works~\cite{you2019pseudo, wang2019pseudo, weng2019monocular} have approached the problem of simulating lidar measurements from images, which allow networks to better reason about 3D scenes more efficiently.

With a similar motivation to our work, Iacono et al.~\cite{iacono2018towards} and Zanardi et al.~\cite{zanardi2019cross} address the issue of transferring learning from images to events by running a network trained on images on the grayscale images produced by some event cameras such as the DAVIS~\cite{brandli2014240}, and using these outputs as ground truth to train a similar network for events. However, these methods treat the frame based outputs as ground truth, and so will learn biases and mistakes made by the frame based network (e.g. the best mAP of the grayscale network in Zanardi et al.~\cite{zanardi2019cross} is 0.59, resulting in a mAP for the event based network of 0.26).

As an alternative approach, our work follows the philosophy of using GANs for image to event simulation. We then use the simulated events to train directly on the ground truth labels for the corresponding images, which should be at least as accurate if not better than outputs from a frame based network trained on these labels.

%% file: tex/method.tex
\section{Method}
The generative portion of our pipeline consists of a U-Net~\cite{ronneberger2015u} encoder-decoder network, as used in Zhu et al.~\cite{zhu2019unsupervised} and Rebecq et al.~\cite{rebecq2019events}. The generator takes as input a pair of grayscale images, concatenated along the channel dimension, and outputs a volumetric representation of the events, described in Section~\ref{sec:representation}. To constrain this output, we apply an adversarial loss, described in Section~\ref{sec:adversarial}, as well as a pair of cycle consistency losses, described in Section~\ref{sec:cycle}. The full pipeline for our method can be found in Figure~\ref{fig:pipeline}.

\subsection{Event Representation}
\label{sec:representation}
The most compact way to represent a set of events is as a set of 4-tuples, consisting of the $x,y$ position, timestamp, $t$, and polarity, $p$. However, regressing points in general is a difficult task, and faces challenges such as varying numbers of events and permutation invariance. 

In this work, we bypass this issue by instead regressing an intermediate representation of the events as proposed by Zhu et al.~\cite{zhu2019unsupervised}. In this representation, the events are scattered into a fixed size 3D spatiotemporal volume, where each event, $(x, y, t, p)$ is inserted into the volume, which has $B=9$ temporal channels, with a linear kernel:
\begin{align}
t^*_i =& (B-1)(t_i - t_1) / (t_{N} - t_1)\\
V(x,y,t)=&\sum_{i} \max(0, 1-|t-t^*_i|)
\end{align}
This retains the distribution of the events in x-y-t space, and has shown success in a number of tasks ~\cite{zhu2019unsupervised, chaney2019learning,rebecq2019events}. 

However, we deviate from the prior work in that we generate separate volumes for each polarity, and concatenate them along the time dimension. This results in a volume which is strictly non-negative, allowing for a ReLU as the final activation of the network, such that the sparsity in the volume is easily preserved. 

In addition, we normalize this volume similar to Rebecq et al.~\cite{rebecq2019events}, with an additional clipping step, as follows:
\begin{align}
\hat{V}(x, y, t) =& \frac{\min(V(x, y, t), \eta_{98})}{\eta_{98}}
\end{align}
where $\eta_{98}$ is the 98th percentile value in the set of non-zero values of $V$. This equates to a clipping operation, followed by a normalization such that the volume lies in $[0, 1]$. The clipping is designed to reduce the effect of hot pixels, which have an erroneously low contrast thresholds and thus generate a disproportionately many events, skewing the range.
\input{tex/loss_ablation_fig.tex}
\subsection{Adversarial Loss}
\label{sec:adversarial}
Perhaps the most direct way to supervise this network is to apply a direct numerical error, such as a L1 or L2 loss, between the predicted and real events. However, given a pair of images, the number of plausible event distributions between the images is extremely large (two images can not constrain the exact motion in between them). Such a direct loss would likely cause the network to overfit to the trajectories observed in the training set and fail to generalize. 

Instead, we apply an adversarial loss~\cite{goodfellow2014generative}. This loss simply constrains the generated events to follow the same distribution as the real ones, and avoids directly constraining the network to memorizing the trajectories seen at training time. For each event-image pair, $(x, y)$, we regress a generated event volume using our network, $G$, and then pass the generated events and real events through a discriminator network, $D$, which predicts the probability that its input is from real data. Our discriminator is a 4 layer PatchGAN classifier~\cite{isola2017image}. We alternatingly train the generator and discriminator, with the discriminator trained 2 steps for every 1 of the generator, using the hinge adversarial loss~\cite{lim2017geometric, tran2017hierarchical}:
\begin{align}
\loss_D=&-\mathbb{E}_{(x,y)\sim p_{\text{data}}}[\min(0, -1+D(x, y))]\nonumber\\
&-\mathbb{E}_{y\sim p_{\text{data}}}[\min(0, -1-D(G(y), y))]\\
\loss_G=&-\mathbb{E}_{y\sim p_{\text{data}}}D(G(y), y)
\end{align}

\subsection{Cycle Consistency Losses}
\label{sec:cycle}
However, GANs are typically difficult to train, especially with a high dimensional output space such as an event volume. In addition, there are no guarantees on the simulated events retaining the salient information in the images, such as accurate motion and intensity information.

To this end, we apply an additional pair of losses which constrain the generated events to encode this motion and intensity information. In particular, we pre-train a pair of networks for optical flow estimation and image reconstruction from real events, using the pipeline in EV-FlowNet~\cite{zhu2018ev}. 

The flow network takes as input the event volume, and outputs a per pixel optical flow. Supervision is applied by warping the previous image to the time of the next image using the predicted flow, and applying an L1 loss between the warped and original image, as well as a local smoothness constraint. 

The image reconstruction network takes as input the previous image and the event volume, and outputs the predicted next image, and is directly supervised by a L1 loss between the reconstructed and original image. The previous image is provided as input as we found that the image reconstruction network tended to overfit to the training set without it. Prior work by Rebecq et al.~\cite{rebecq2019events} has circumvented this by training in a recurrent fashion, but doing so would require multiple passes through the recurrent network, which is undesirably expensive when the goal is to train the generator network. In addition, we summarize the event volume by summing along the time dimension. This is to maintain the invariance to permutation across time of the events. For example, two events occurring at the start of the window vs. two events at the end of the window should generate the same output image. The input, then, to the reconstruction network, is a 2-channel image consisting of the previous image and the summed event volume.

In summary, the cycle consistency losses are:
\begin{align}
\loss_{F}=&\sum_{\x}\|I_0(\x-F(\x; G))-I_1(\x)\|_1 \nonumber\\
& + \lambda_1 \left(\left\|\frac{dF}{dx}(\x; G)\right\|_1+ \left\|\frac{dF}{dy}(\x; G)\right\|_1\right)\\
\loss_{R}=&\sum_{\x}\|\hat{I}_1(\x; G, I_0)-I_1(\x)\|_1\\
\loss_{\text{cycle}}=&\loss_F+\loss_G
\end{align}

When training the generator network, we pass the output from the generator as input to each of the pre-trained networks, and apply the same losses used to train each. However, in this case, we freeze the weights of each pre-trained network, such that the generator must tune its output to generate the best input for each pre-trained network. Both cycle consistency networks share the same architecture as the generator network, with the losses applied each time the generator is updated in the adversarial framework. The final losses at each step are:
\begin{align}
&\text{Generator step:}
&\loss_{GS}&=\loss_G+\loss_\text{cycle}\label{eq:gen}\\
&\text{Discriminator step:}
&\loss_{DS}&=\loss_D\label{eq:dis}
\end{align}

These losses provide useful gradients early in training, when the adversarial loss is typically unstable, and embed motion and appearance information in the predicted event volumes. Figure~\ref{fig:loss_ablation} shows the effect of each loss on the output of the generator. 

In summary, the adversarial loss enforces sparsity in the event volume and similarity between the fake and real event distributions. The flow loss enforces motion information to be present within the volume, while the reconstruction loss enforces regularity in the number of events generated by the same point. This is particularly evident when one visualizes the image of the average timestamp at each pixel, where extremely low (but non-zero) values may be hidden in the count image, and where motion trails are clearly visible.

We also implement the tips prescribed by Gulrajani et al.~\cite{gulrajani2017improved} and Brock et al.~\cite{brock2018large}. In particular, we apply spectral normalization~\cite{miyato2018spectral} in the encoder of the generator, and batch normalization~\cite{ioffe2015batch} for the entire generator, while the discriminator has neither types of normalization. We also add noise to the labels seen by the discriminator by randomly flipping the labels from real to fake 10\% of the time, as recommended by Chintala et al.~\cite{chintala2016train}.

%% file: tex/loss_ablation_fig.tex
\begin{figure*}[t!]
    \centering
    \begin{subfigure}[b]{0.28\linewidth}
    \includegraphics[width=\textwidth]{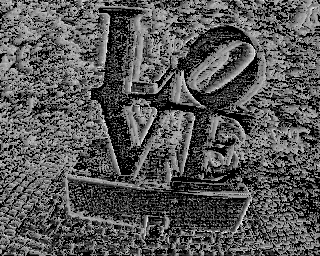}
    \caption{l1-flow-recons}
    \end{subfigure}
    \begin{subfigure}[b]{0.28\linewidth}
    \includegraphics[width=\textwidth]{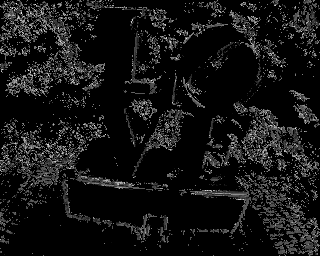}
    \caption{adv.}
    \end{subfigure}
    \begin{subfigure}[b]{0.28\linewidth}
    \includegraphics[width=\textwidth]{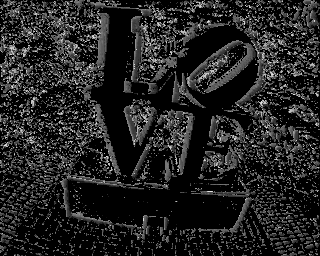}   
    \caption{adv.-recons}
    \end{subfigure}

    \vspace{1pt}
    \begin{subfigure}[b]{0.28\linewidth}
    \includegraphics[width=\textwidth]{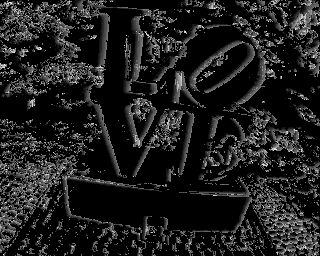}   
    \caption{adv.-flow}
    \end{subfigure}
    \begin{subfigure}[b]{0.28\linewidth}
    \includegraphics[width=\textwidth]{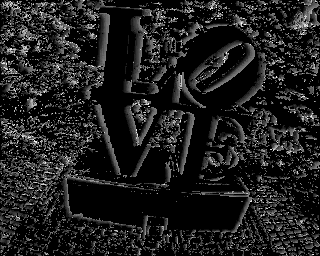}
    \caption{adv-flow-rec}
    \end{subfigure}
    \begin{subfigure}[b]{0.28\linewidth}
    \includegraphics[width=\textwidth]{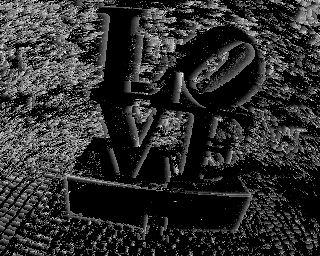}
    \caption{real}
    \end{subfigure}
    \caption{Outputs from models trained with subsets of our proposed loss, all models trained with the same hyperparameters. Events are visualized as average timestamp images, i.e. the average timestamp at each pixel. Any voxel with non zero value will generate a color in the average timestamp image, allowing us to see the sparsity of the volume. \textbf{(a)}: L1 reconstruction loss in place of the adversarial loss, causing artifacts in the events, and no sparsity achieved, as observed in the interior of the `LOVE' symbol in the time image. \textbf{(b)}: Adversarial loss only. Model struggles to converge, and requires significant hyperparameter tuning in order to achieve good results. \textbf{(c)}: Adversarial loss and reconstruction loss. Model is now stable, but the events do not have motion information. The image should have a gradient in the motion direction. \textbf{(d)}: Adversarial loss and flow loss. Motion direction can now be seen in the time image, but events are not generated in many areas. \textbf{(e)}: Adversarial loss, flow and reconstruction losses. Motion trails can now be clearly seen in the time image (see letters). \textbf{(f)}: Real events. Note that our method typically underestimates the amount of motion in the scene.}
    \label{fig:loss_ablation}
\end{figure*}

%% file: tex/gan_outputs.tex
\begin{figure*}[t!]
\centering
    \includegraphics[width=0.32\linewidth]{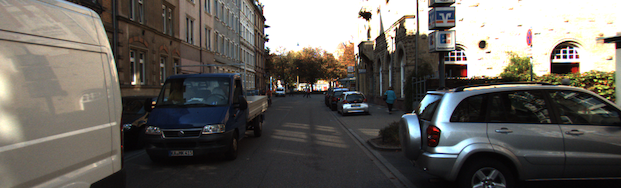}\hfil%
    \includegraphics[width=0.32\linewidth]{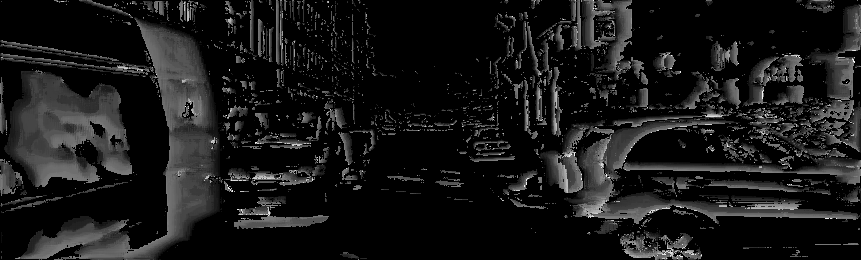}\hfil%
    \includegraphics[width=0.32\linewidth]{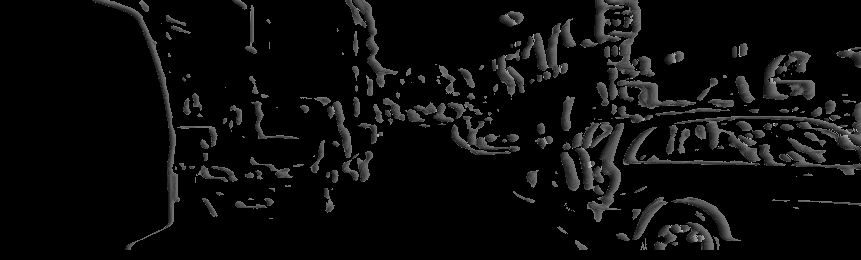}
    
    \vspace{1pt}
    
    \includegraphics[width=0.32\linewidth]{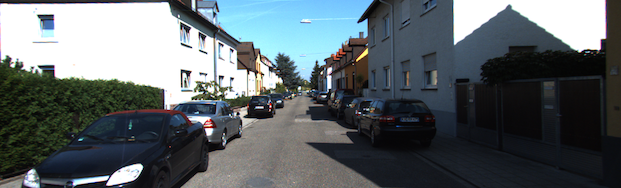}\hfil%
    \includegraphics[width=0.32\linewidth]{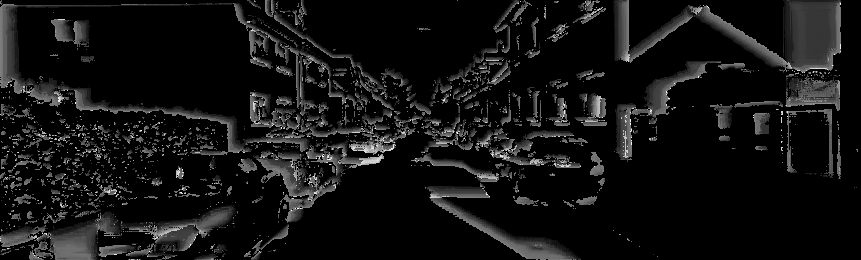}\hfil%
    \includegraphics[width=0.32\linewidth]{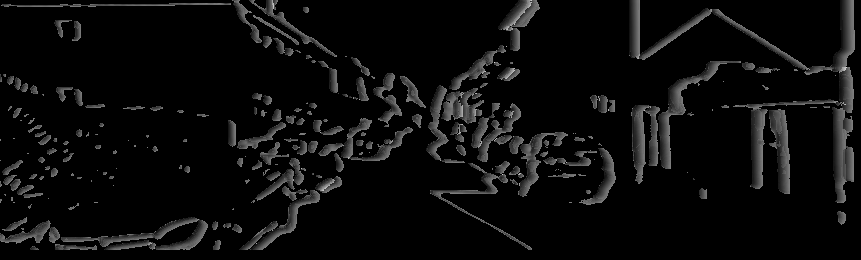}
    
    \vspace{1pt}
    
    \includegraphics[width=0.32\linewidth]{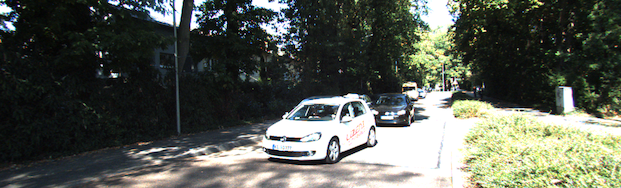}\hfil%
    \includegraphics[width=0.32\linewidth]{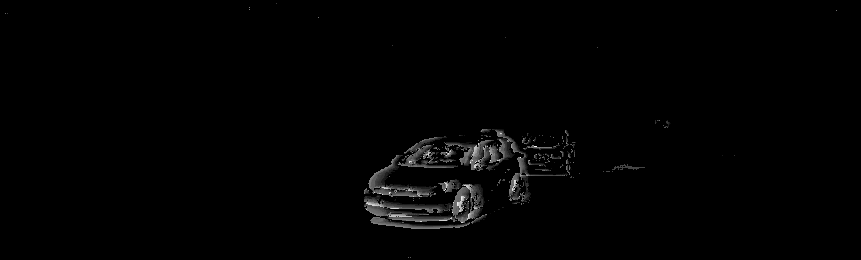}\hfil%
    \includegraphics[width=0.32\linewidth]{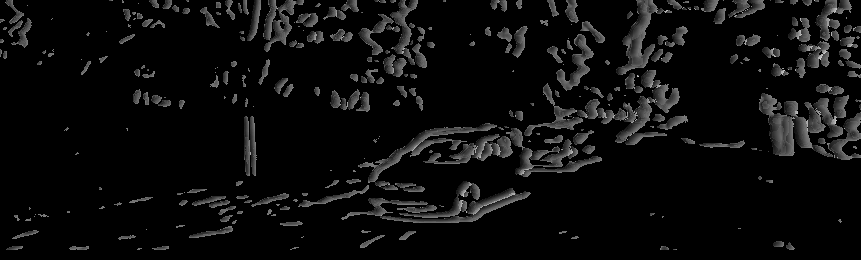}

    \vspace{1pt}    
    
    \includegraphics[width=0.26\linewidth]{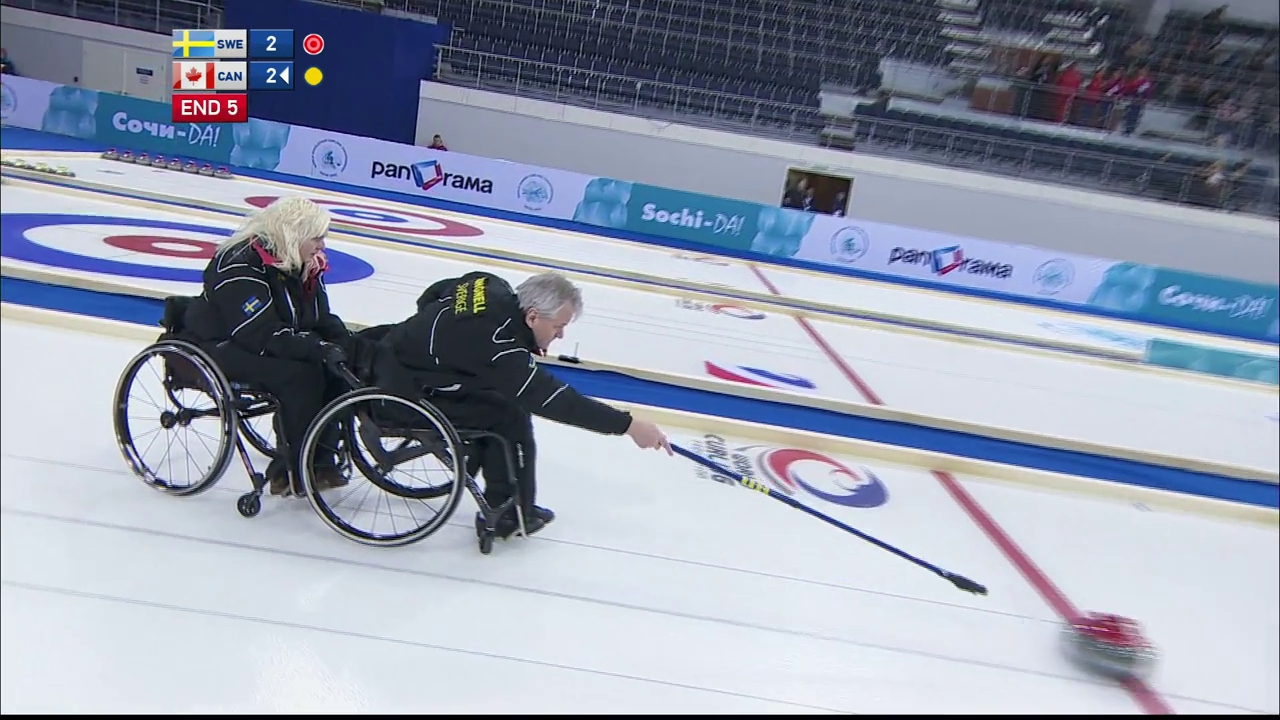}\hfil%
    \includegraphics[width=0.26\linewidth]{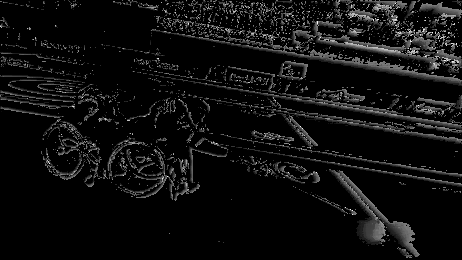}\hfil%
   \includegraphics[width=0.26\linewidth]{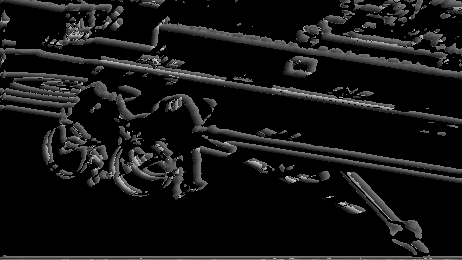}
    \vspace{1pt}    
    
    \includegraphics[width=0.26\linewidth]{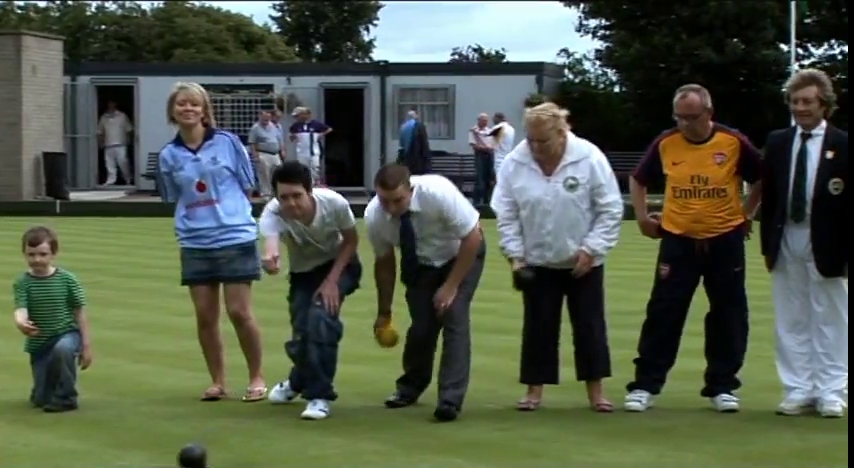}\hfil%
    \includegraphics[width=0.26\linewidth]{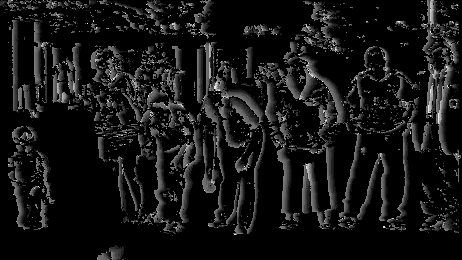}\hfil%
    \includegraphics[width=0.26\linewidth]{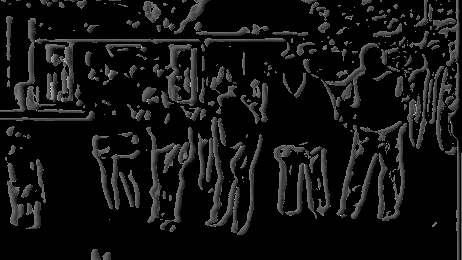}
    
    \vspace{1pt}    
    \begin{subfigure}[b]{0.26\textwidth}
    \includegraphics[width=\textwidth]{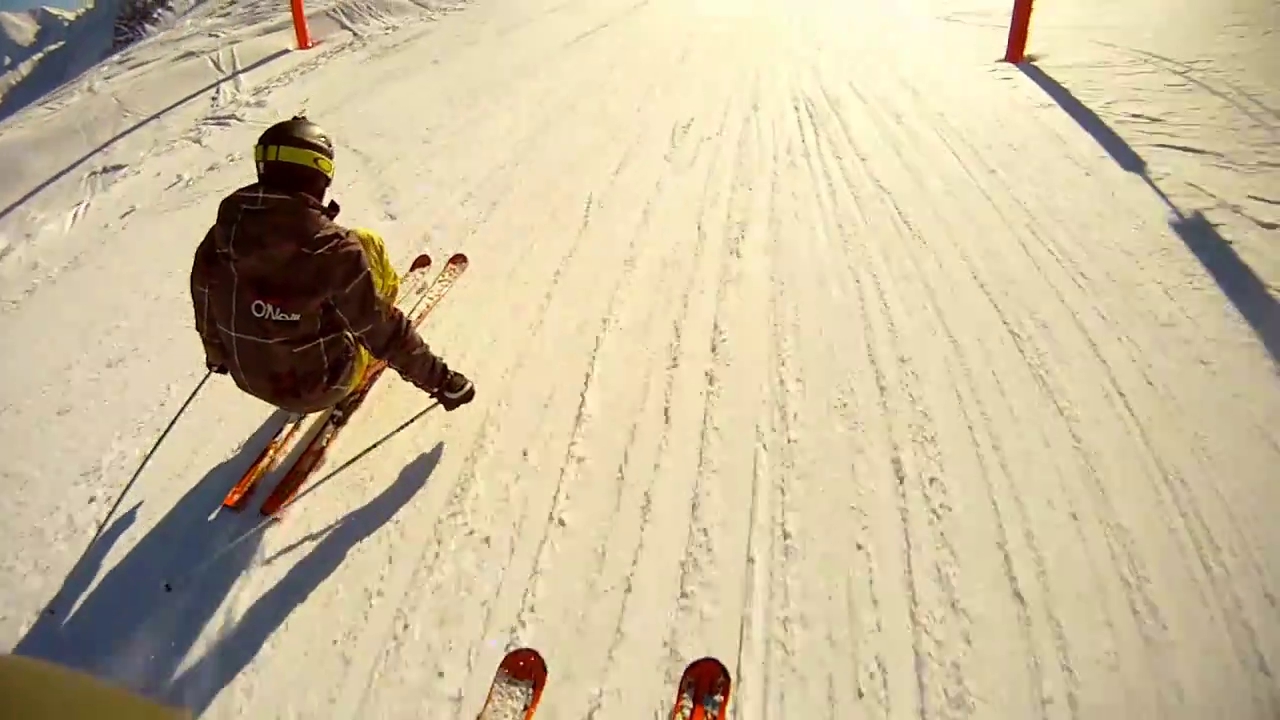}
    \caption{Input Frame}
    \end{subfigure}\hfil%
    \begin{subfigure}[b]{0.26\textwidth}
    \includegraphics[width=\textwidth]{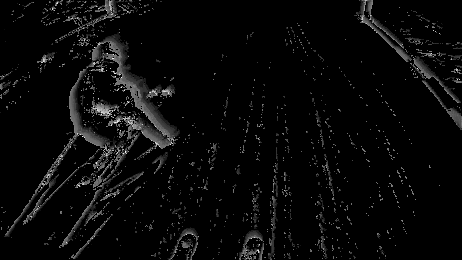}\hfil%
    \caption{EventGAN}
    \end{subfigure}\hfil%
    \begin{subfigure}[b]{0.26\textwidth}    
    \includegraphics[width=\textwidth]{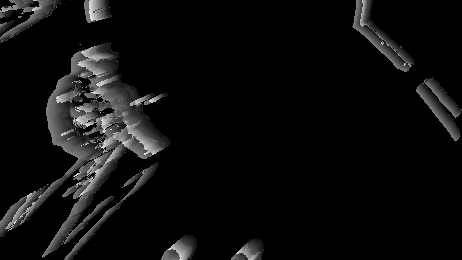}
    \caption{ESIM}
    \end{subfigure}\hfil%
    \caption{Sample outputs generated by EventGAN, compared to ESIM~\cite{rebecq2019events}, visualized as images of the average timestamp at each pixel. Top images are from the KITTI dataset~\cite{Geiger2012CVPR}, bottom are from MPII~\cite{andriluka14cvpr}. Compared to ESIM, our method is able to more accurately capture the motion in the scene, and capture fine grain information.}
    \label{fig:gan_outputs}
\end{figure*}

%% file: tex/detection_qualitative.tex
\begin{figure*}[t!]
    \centering
    EventGAN\\
    \vspace{1pt}    
    \includegraphics[width=0.22\linewidth]{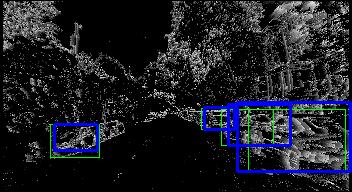}
    \includegraphics[width=0.22\linewidth]{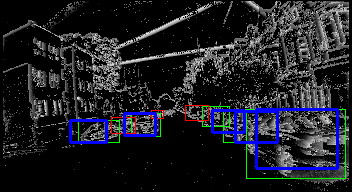}
    \includegraphics[width=0.22\linewidth]{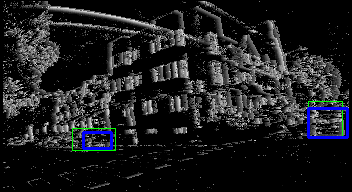}
    \includegraphics[width=0.22\linewidth]{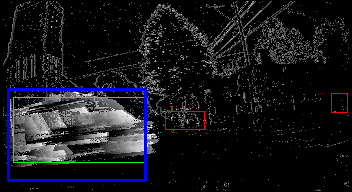}
    
    \vspace{1pt}
    ESIM\\
    \vspace{1pt}
    \includegraphics[width=0.22\linewidth]{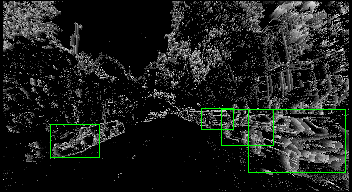}
    \includegraphics[width=0.22\linewidth]{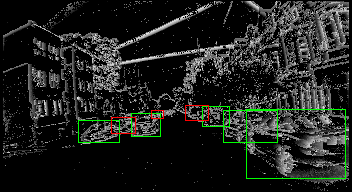}
    \includegraphics[width=0.22\linewidth]{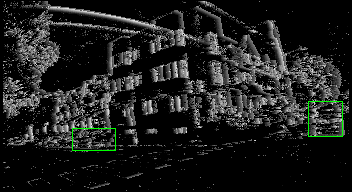}
    \includegraphics[width=0.22\linewidth]{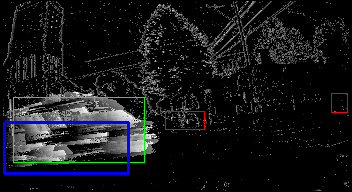}
    
    \vspace{1pt}    
    Frame\\
    \vspace{1pt}
    \includegraphics[width=0.22\linewidth]{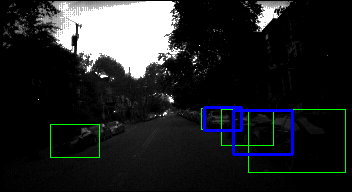}
    \includegraphics[width=0.22\linewidth]{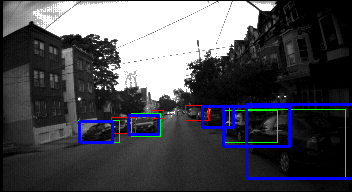}
    \includegraphics[width=0.22\linewidth]{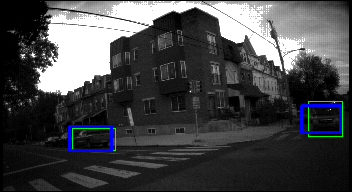}
    \includegraphics[width=0.22\linewidth]{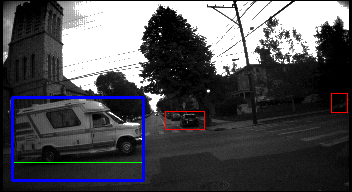}
    \caption{Selected qualitative results of our car detection pipeline using the YOLOv3 network~\cite{redmon2018yolov3}. Detections are in blue, GT labels in green, and don't care regions in red. For explanation of the methods, please see Table~\ref{tab:detection_results}.}
    \label{fig:detection_qualitative}
    
\end{figure*}

%% file: tex/experiments.tex
\section{Experiments}
We train our network using the RAdam~\cite{liu2019variance} optimizer for 100 epochs on events and images from the indoor$\_$flying and outdoor$\_$day sequences in the MVSEC dataset~\cite{zhu2018multivehicle}, as well as a newly collected dataset consisting of recordings from a DAVIS-346b camera~\cite{brandli2014240}, consisting of short ($<$60s) sequences with a number of different scenes and motions, in order to capture a large range of event distributions. As the objective of this work is to produce an event simulator which operates well on existing image datasets, we did not train on scenes which are challenging for images (e.g. night time driving). In total, the training set consists around 30 mins of data. During training, we perform weighted sampling from this dataset, with a 80\%/20\% split between the new data and MVSEC. Each input to the network consists of a pair of images, randomly picked between 1 and 6 frames apart, and the events between them.

Quantitative evaluations of generative models is difficult, as measuring how well the predicted events fit the true event distribution requires knowledge of the true event distribution. For images, networks trained a large corpus of image data are used to model these distributions, and metrics such as the Inception Score~\cite{salimans2016improved} or the Fr\' echet Inception Distance~\cite{heusel2017gans} are applied using these networks. However, this results in a second chicken and egg problem, as no such corpus of event data currently exists. 

Instead, we evaluate our method directly on a set of downstream tasks, and demonstrate that our simulated events are able to train networks for complex tasks which generalize to data with real events. In Sections~\ref{sec:human_pose} and~\ref{sec:detection} we describe our experiments for 2D human pose estimation and object detection, respectively.

\subsection{2D Human Pose Estimation}
\label{sec:human_pose}
We train a 2D human pose detector for events based on the publicly available code from Xiao et al.~\cite{xiao2018simple}, which uses an encoder-decoder style network to regress a heatmap for each desired joint. We use a ResNet-50~\cite{he2016deep} encoder, pretrained on ImageNet~\cite{ILSVRC15}. For event inputs, we modify the number of input channels in the first layer, and randomly initialize the weights of this layer. The network is then trained on a 80\%/20\% split between the MPII~\cite{andriluka14cvpr} and Human3.6M~\cite{h36m_pami} datasets. For each ground truth pose, the pair of images either 1 or 2 frames before and after the target frame are selected at random, and passed into the generator network to generate a simulated event volume.

We evaluate our method on the DHP19~\cite{calabrese2019dhp19} dataset, which consists of 3D joint positions of a human subject, recorded with motion capture, with events from four cameras surrounding the subject. Using the camera calibrations, we project these 3D joint positions into 2D image positions for each camera. Following the experiment schedule by Calabrese et al.~\cite{calabrese2019dhp19}, we use as a test set data from subjects 13-17 and cameras 2-3. As our method does not include any temporal consistency, we remove sequences with hand motions only, where most of the body is static and does not generate any events. This results in 16 motions across 5 subjects and 2 cameras. Following Calabrese et al.~\cite{calabrese2019dhp19}, we divide each sequence into chunks of 7500 events per camera, and evaluate on the average pose within each window.
\input{tex/detection_table.tex}
One issue with this direct evaluation is that the marker positions for DHP19 vary significantly from those in MPII and H36M. In order to overcome this offset between the joint positions, we freeze all but the final linear layer of our network, and fine tune this layer on the DHP19 training set (subjects 1-12, cameras 2-3). This is equivalent to training a linear model on the activations from the second to last layer, as is common in the self-supervised learning literature~\cite{goyal2019scaling}.

\subsection{Object Detection}
\label{sec:detection}
We train a detection network using the YOLOv3 pipeline~\cite{redmon2018yolov3}. We initialize the network from a pretrained YOLOv3 network with spatial pyramid pooling, with the first input layer randomly initialized. The network is trained on simulated events from the KITTI Object Detection dataset~\cite{Geiger2012CVPR}, with the target frame and either the frame one or two frames prior. 

\input{tex/human_pose_figs.tex}
\input{tex/pose_table.tex}

\subsection{The Event Car Detection Dataset}
\label{sec:dataset}
For evaluation, we generated a novel dataset for car bounding box annotations for event data. Our dataset consists of 250 labeled images from the MVSEC~\cite{zhu2018multivehicle} outdoor driving dataset, with corresponding timestamps. For each image, raters label bounding boxes for all cars within the scene, while also separating the cars into easy (large, no occlusion), hard (medium, or partial occlusion) or don't care (mostly occluded or too small) categories. In total, there are 451 easy instances, 506 hard instances and 959 don't care instances. This dataset will be publicly available.

\subsection{Competing Methods}
We additionally simulate the MPII, H36M and KITTI datasets using ESIM~\cite{rebecq2018esim}, by simulating a random affine transform of each image in the dataset, similar to the method used by Rebecq et al.~\cite{rebecq2019events}. Using this simulated data, we train the same networks described in Sections~\ref{sec:human_pose} and \ref{sec:detection}. For both experiments, we also train networks on real data as a baseline. For object detection, we train a network on the grayscale frames from KITTI, and evaluate on the grayscale frames from MVSEC and DDD17. For human pose estimation, we train a network on the events in the training set (subjects 1-12) of DHP19.

%% file: tex/detection_table.tex
\begin{table*}[h!]
\begin{center}
\begin{tabular}{ c | c | c | c | c | c | c}
Training Data & Precision & Easy recall & Hard recall & Comb recall & AP & F-1\\
\hline
EventGAN & 0.42 & \textbf{0.57} & \textbf{0.34} & \textbf{0.45} & \textbf{0.30} & 0.44\\
ESIM & 0.23 & 0.08 & 0.02 & 0.05 & 0.02 & 0.09\\
Frame & \textbf{0.57} & 0.48 & 0.27 & 0.37 & 0.29 & \textbf{0.45}\\
\end{tabular}
\end{center}
\caption{Object detection results on the Event Car Detection dataset. Metrics adopted from the PASCAL VOC challenge~\cite{everingham2010pascal}. The EventGAN and ESIM models are trained on simulated events from the KITTI dataset, while the Frame model is trained on the real image frames from the KITTI dataset.}
\label{tab:detection_results}
\end{table*}

%% file: tex/human_pose_figs.tex
\begin{figure*}[h]

 \begin{subfigure}[b]{0.49\textwidth}
     \centering
  \includegraphics[width=0.24\linewidth]{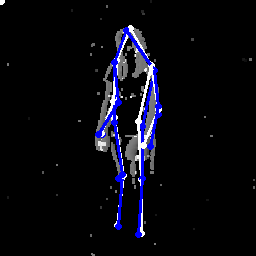}
    \includegraphics[width=0.24\linewidth]{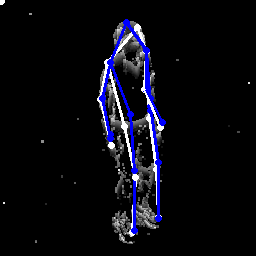}
    \includegraphics[width=0.24\linewidth]{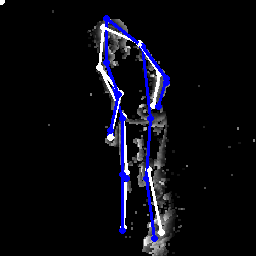}
    \includegraphics[width=0.24\linewidth]{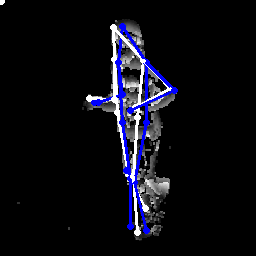}
    \caption*{Real Events}
    \end{subfigure}
 \begin{subfigure}[b]{0.49\textwidth}    
     \centering
    \includegraphics[width=0.24\linewidth]{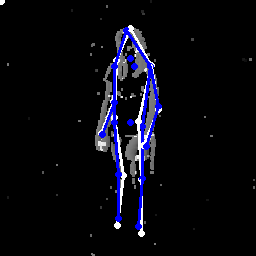}
    \includegraphics[width=0.24\linewidth]{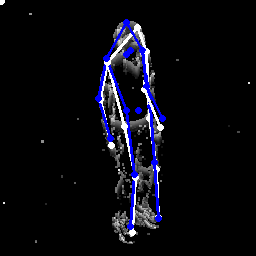}
    \includegraphics[width=0.24\linewidth]{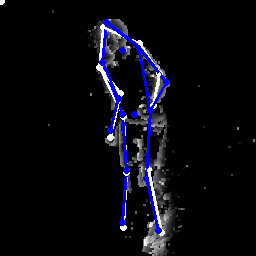}
    \includegraphics[width=0.24\linewidth]{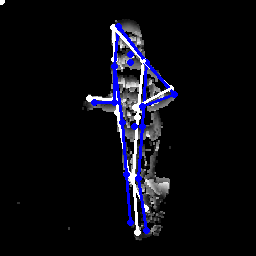}
    \caption*{EventGAN-fine-30}
 \end{subfigure}
     \begin{subfigure}[b]{0.49\textwidth}
         \centering
    \includegraphics[width=0.24\linewidth]{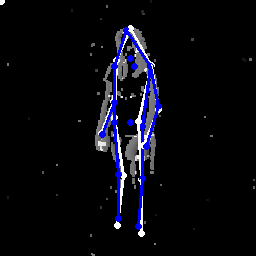}
    \includegraphics[width=0.24\linewidth]{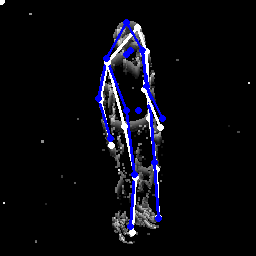}
    \includegraphics[width=0.24\linewidth]{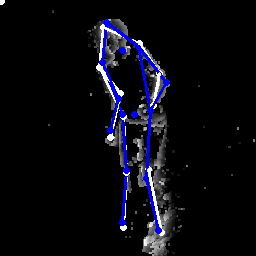}
    \includegraphics[width=0.24\linewidth]{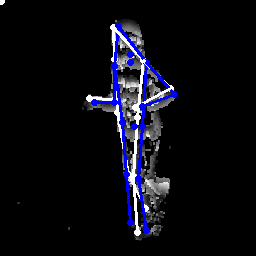}
    \caption*{ESIM-fine-30}
    \end{subfigure}
     \begin{subfigure}[b]{0.49\textwidth}
         \centering
    \includegraphics[width=0.24\linewidth,trim={4cm, 0, 79pt, 0},clip]{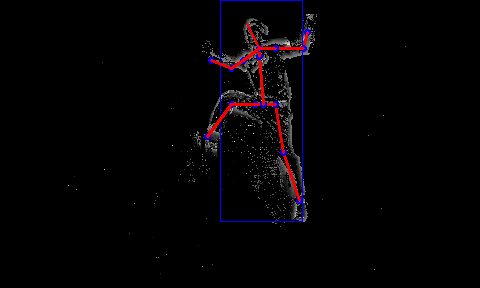}
    \includegraphics[width=0.24\linewidth,trim={4cm, 0, 79pt, 0},clip]{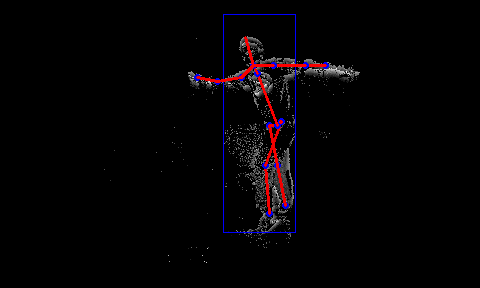}
    \includegraphics[width=0.24\linewidth,trim={4cm, 0, 79pt, 0},clip]{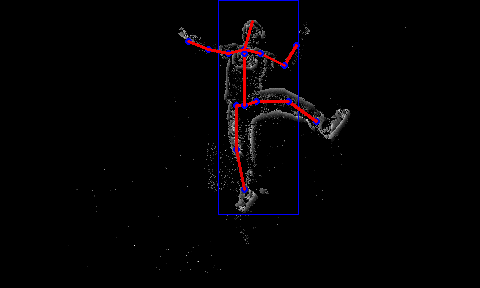}
    \includegraphics[width=0.24\linewidth,trim={4cm, 0, 79pt, 0},clip]{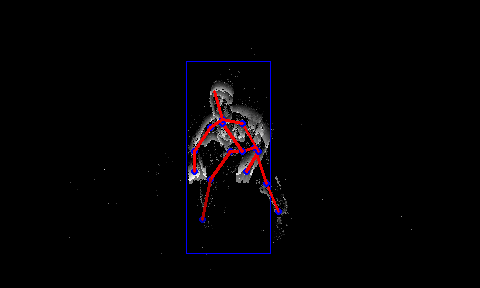}
    \caption*{EventGAN Evaluated on Custom Data}
    \end{subfigure}
    \caption{Qualitative results of our human pose estimation on real event data. The first three sets are evaluated on samples from the DHP19 dataset~\cite{calabrese2019dhp19}, where ground truth is in white and predictions are in blue. Our model is able to achieve accuracy on par with a model directly trained on the real data after 30 epoch of fine tuning only the last linear layer. The last set shows our YOLOv3 detection pipeline combined with our human pose estimator. The detection network is trained on MPII to detect the human in the scene (blue box), which is fed into the human pose estimator to estimate the 2D joint positions (MPII format). Best viewed in color.}
    \label{fig:human_pose_pred}
\end{figure*}

%% file: tex/pose_table.tex
\begin{table*}[h!]
\begin{center}
\begin{tabular}{ c | c  c | c  c  c | c  c  c | c}
 & \multicolumn{2}{c|}{Pretrained only} & \multicolumn{3}{c|}{1 Epoch} & \multicolumn{3}{c|}{30 Epochs} & 140 epochs\\

 & EventGAN & ESIM & EventGAN & ESIM & Real & EventGAN & ESIM & Real & Real\\
\hline
MPJPE $\downarrow$ & \textbf{14.55} & 19.57 & \textbf{6.76} & 7.58 & 8.94 & \textbf{6.44} & 6.54 & 6.75 & \textbf{6.39}\\
PCKh@50 $\uparrow$ & \textbf{45.47} & 40.53 & \textbf{87.70} & 85.89 & 80.55 & \textbf{90.19} & 89.93 & 87.53 & 89.86
\end{tabular}
\end{center}
\caption{Human pose estimation results in MPJPE (pix.) (lower is better) and PCKh$@$50 (higher is better). All EventGAN and ESIM models are first pretrained on simulated events from the MPII and H36M datasets, and then the final linear layer is fine tuned on the DHP19 training set for the specified number of epochs. The Real models are trained directly (whole model) on the DHP19 training set for the specified number of epochs.}
\label{tab:pose_results}
\end{table*}

%% file: tex/results.tex
\section{Results}
\subsection{2D Human Pose Estimation}
We evaluate our method on the mean per joint position error (MPJPE)~\cite{calabrese2019dhp19}, $\frac{1}{N}\sum_i^N\|x_i-\hat{x}_i\|_2$, as well as PCKh$@$50 (percentage of correct keypoints)~\cite{andriluka14cvpr}, which measures the percentage of joint predictions with error less than $50\%$ of the head size. We define head size as 0.6$\times$ the distance between the head and the midpoint between the shoulders. 

In Table~\ref{tab:pose_results}, we compare a network trained on simulated events from EventGAN, ESIM, and a network trained directly on the DHP19 training set. We also report results from fine tuning the final linear layer of the network on the DHP19 training set for both EventGAN and ESIM. Qualitative results from both DHP19 and out of sample data can be found in Figure~\ref{fig:human_pose_pred}. From these results, we can see that the data generated by EventGAN is able to train a network to learn representations that are very close to the true data. After only one epoch of fine tuning, and only of the final layer, we are able to achieve significantly higher accuracy than training on the real data, and come close to the accuracy of a network trained for 140 epochs on real data. However, the gap between ESIM and our method is also relatively small. This is largely due to the low difficulty of the dataset, as even training on real events converges to a relatively good solution after only one epoch of training. This was observed even when testing with much smaller networks, although they converge to a lower accuracy. The dataset is also much cleaner, and as such is closer to the ESIM outputs.

\subsection{Object Detection}
We evaluate our method according to the precision-recall statistics defined by the PASCAL VOC challenge~\cite{everingham2010pascal}. Predictions with confidence $<0.2$ are removed, and non-maximum suppression is applied for boxes with IoU $> 0.2$. In total, we report precision, recall on the easy and hard classes, as well as the AP and F-1 scores for each training input in Table~\ref{tab:detection_results}. We provide qualitative results in Figure~\ref{fig:detection_qualitative}.

From these results, we observed that our method is able to achieve reasonably strong results, and comes close to matching the performance of the network with frame inputs, which was trained on real data. The difference in performance implies a small sim-to-real gap, but may also simply be due to a stronger signal in the images for certain frames (although this may also be true the other way round). On the other hand, the sim-to-real gap is significant when training on ESIM. As the true event distribution differs largely from the simulated data, the network is only able to perform accurate detections when the input has relatively low noise (e.g. Figure~\ref{fig:detection_qualitative} right), resulting in very low recall.

%% file: tex/conclusions.tex
\section{Conclusions}
In this work, we have proposed a novel method for training supervised neural networks for events using image data by way of image to event simulation. Given events and images from an event camera, our deep learning pipeline is able to accurately simulate events from a pair of grayscale images from existing image datasets. These events can be used to train downstream networks for complex tasks such as object detection and 2D human pose estimation, and generalize to real events.

The largest limitation of this work is the need for a pair of frames (video), thus prohibiting the use of larger image datasets such as ImageNet~\cite{ILSVRC15} and COCO~\cite{lin2014microsoft}. While it is possible to train a GAN to predict events from a single image, this would become a complex future prediction task, as the GAN must hallucinate the motion within the image. Other promising future directions include exploring other event representations, more complicated adversarial architectures, and exploring more complex downstream tasks.